\documentclass[conference]{IEEEtran}
\IEEEoverridecommandlockouts

\usepackage{cite}
\usepackage{amsmath,amssymb,amsfonts}
\usepackage{graphicx}
\usepackage{textcomp}
\usepackage{xcolor}
\usepackage{url}
\usepackage{hyperref}

\usepackage{multirow} 
\usepackage{booktabs} 
\usepackage{array}    
\usepackage{arydshln} 
\usepackage{algorithm}
\usepackage{algpseudocode}

\hypersetup{
    colorlinks=true,
    linkcolor=blue,
    filecolor=blue,      
    urlcolor=magenta,
    citecolor=magenta,
}
\def\BibTeX{{\rm B\kern-.05em{\sc i\kern-.025em b}\kern-.08em
    T\kern-.1667em\lower.7ex\hbox{E}\kern-.125emX}}
\begin{document}

\title{Universal Domain Adaptation Benchmark for Time Series Data Representation\\
\thanks{$^\star$The authors acknowledge the support of the French Agence Nationale de la Recherche (ANR), under grant ANR-23-CE23-0004 (project ODD) and of the STIC AmSud project DD-AnDet under grant N°51743NC.}
}


\author{\IEEEauthorblockN{Romain Mussard, Fannia Pacheco, Maxime Berar, Gilles Gasso and Paul Honeine}
\IEEEauthorblockA{\textit{Univ Rouen Normandie, INSA Rouen Normandie, Normandie Univ, LITIS UR 4108}\\
\textit{F-76000 Rouen, France} \\
Email: romain.mussard@univ-rouen.fr}
}

\maketitle

\begin{abstract}

Deep learning models have significantly improved the ability to detect novelties in time series (TS) data. This success is attributed to their strong representation capabilities. However, due to the inherent variability in TS data, these models often struggle with generalization and robustness. To address this, a common approach is to perform Unsupervised Domain Adaptation, particularly Universal Domain Adaptation (UniDA), to handle domain shifts and emerging novel classes. While extensively studied in computer vision, UniDA remains underexplored for TS data. This work provides a comprehensive implementation and comparison of state-of-the-art TS backbones in a UniDA framework. We propose a reliable protocol to evaluate their robustness and generalization across different domains. The goal is to provide practitioners with a framework that can be easily extended to incorporate future advancements in UniDA and TS architectures. Our results highlight the critical influence of backbone selection in UniDA performance and enable a robustness analysis across various datasets and architectures.

\end{abstract}

\section{Introduction}

Anomaly detection plays a crucial role in time series (TS) analysis, enabling the detection of irregular patterns or deviations, such as the emergence of new classes \cite{TSsurvey2}. Detection performance has been enhanced by deep learning when TS data is accurately represented in a latent space \cite{Trirat2024UniversalTR}. Contrastive approaches such as TS2VEC \cite{lee2024soft} and softCL \cite{ts2vec} have been designed to improve TS representation. However, they are expensive to train and require large datasets, making them often unsuitable for anomaly and novelty detection, where data is scarce. Simultaneously, new architectures for TS representation learning have enhanced performance not only in detection but also in classification and forecasting \cite{s3_layer, tslanet, ts2vec, fno_li2020fourier}. This is achieved through the integration of TS-specific methodologies such as frequency analysis variants like FNO \cite{fno_li2020fourier}, TSLANet \cite{tslanet}, and S3 \cite{s3_layer}. As in image and text, foundation models for TS have been developed
but face similar problem under distribution shift \cite{liang2024foundation}.

However, practitioners can leverage regular Unsupervised Domain Adaptation (UDA) to fine-tune models for improved generalization across TS datasets \cite{liu2022deep}. UDA attempts to find a domain-invariant feature space for labeled source samples and unlabeled target samples, assuming that both the source and target data share the same classes and number of classes. To detect anomalies — often referred to as Out-Of-Distribution (OOD) samples — in a UDA context, Universal Domain Adaptation (UniDA) was introduced as a more realistic and complex setting, allowing the emergence of OOD samples in the target domain. 
UniDA must simultaneously align the common classes shared across different domains while also accurately detecting OOD samples \cite{UniOT}.

Most UniDA methods are inspired by UDA methods for common sample alignment: UAN \cite{UDA} uses the adversarial approach proposed in \cite{DANN, ADDA}, while PPOT and UniOT leverage Optimal Transport (OT) introduced in \cite{deepJDOT, JUMBOT}. Alternatively, UniAM \cite{UniAM} introduces a novel alignment framework leveraging sparse data representation and Vision Transformer. Unknown sample classification generally relies on thresholding various metrics such as entropy \cite{UDA}, OT masses \cite{UniOT}, or reconstruction error \cite{UniAM}. Such approaches are similar to how anomaly and OOD detection are often performed. OVANet deviates from other approaches and employs a One-vs-All strategy to perform both alignment and unknown sample detection. RAINCOAT \cite{RAINCOAT} is the only one specifically designed for TS, demonstrating how TS-based architectures can improve UniDA performance.

To the best of our knowledge, there is no quantitative evaluation of the contribution of TS-tailored architectures for OOD detection within a UniDA framework. The only benchmark available is ADATIME \cite{ADATIME}, which focuses on evaluating UDA methods for TS classification. However, ADATIME mainly employs CNN-based backbones for its experiments. This work provides a comprehensive implementation and comparison of state-of-the-art TS backbones in a UniDA context. This paper studies six state-of-the-art UniDA methods, each tested in combination with four TS backbones. We propose a reliable protocol to evaluate their robustness and generalization across different domains. The final objective is to provide practitioners in the field with a framework that is easy to use and extend with future advances in UniDA and TS architectures.

The remainder of this paper is organized as follows: Section~\ref{II} formalizes TS representation learning and the UniDA framework. Section~\ref{III} describes the proposed protocol and benchmarked methods, backbones, and datasets. Section~\ref{IV} presents experimental evaluations on TS datasets. Finally, Section~\ref{V} concludes the paper. Our code is available at \url{https://github.com/RomainMsrd/UniDABench}.

\section{Background} \label{II}

\subsection{TS representation learning}

A TS $\mathbf{x}$ is a collection of chronologically ordered data points over time at consistent intervals, $\mathbf{x} = \{ x_1, x_2, \cdots, x_T\}$, where $x_t \in \mathbb{R}^{D}$ with $T$ the TS length and $D$ the number of variables. A TS is said to be univariate if $D = 1$ and multivariate if $D > 1$. A multivariate TS implies the recording of multiple variables at each time step. A TS can exhibit trends, seasonality, and cyclic behaviors. A TS is said to be stationary if its statistical properties (mean, variance) remain constant over time; otherwise, it is non-stationary.

Representation learning over TS consists of learning a model $f : \mathbf{x} \rightarrow \mathbf{z}$, generally called a feature extractor, that maps the TS input $\mathbf{x}$ to a latent space, $\mathbf{z} = \{ z_1, z_2, \cdots, z_R\}$, with $z_r \in \mathbb{R}^{K}$ such that generally $R \leq T$ and $K \leq D$. In some cases, the representation can be point-wise if $T = R$, meaning that each individual time step is represented by a latent vector. Representation learning is associated with increased results on all tasks related to TS: Forecasting, Classification, Anomaly detection and, more rarely, domain adaptation \cite{s3_layer, tslanet, ts2vec, fno_li2020fourier}.

\subsection{Universal Domain Adaptation}

\begin{figure}[t]
    \centering
    \includegraphics[width=\linewidth]{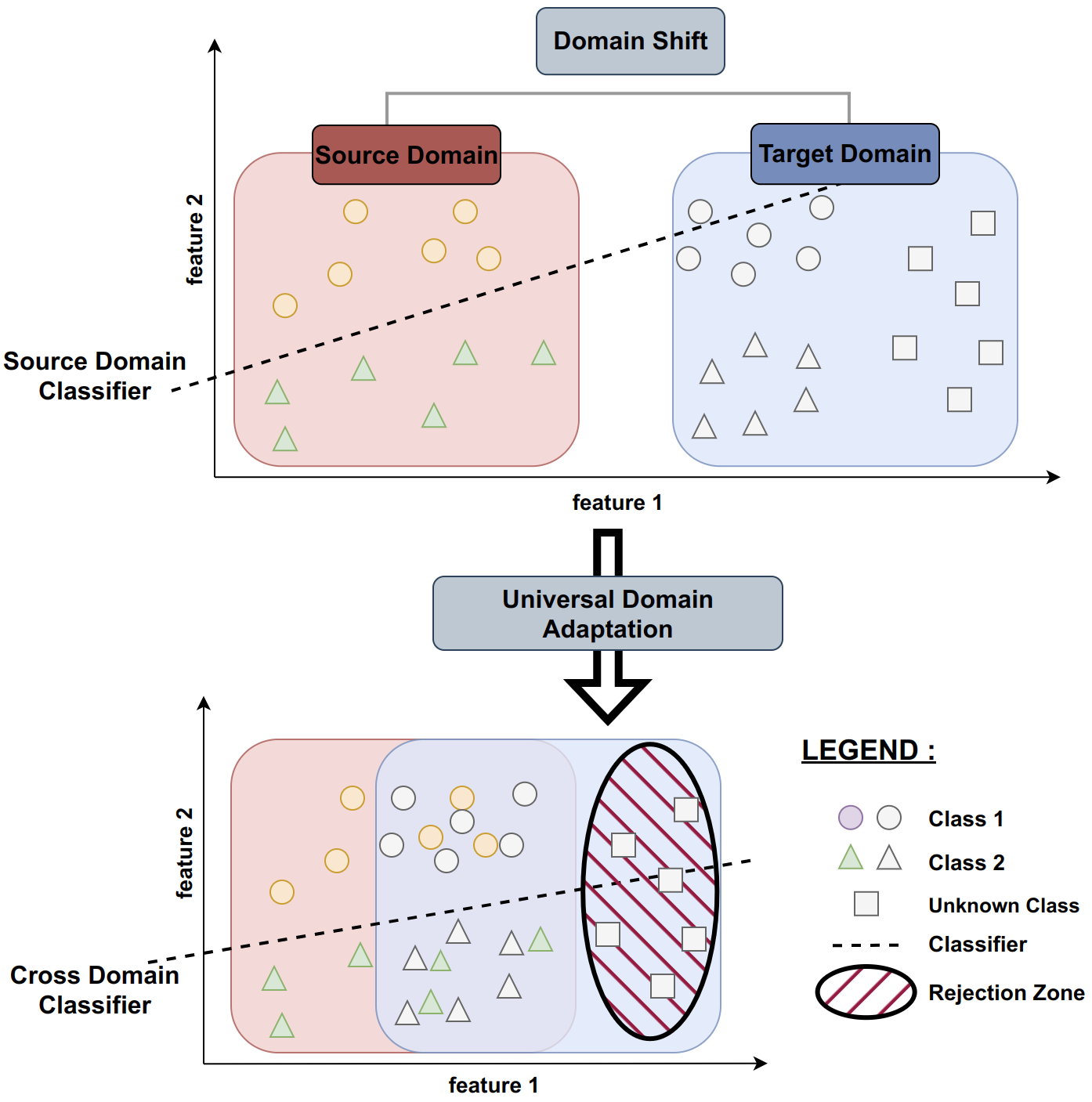}
        \caption{\textbf{Illustration of Universal Domain Adaptation (UniDA).} \textit{Top:} A classifier trained on the source domain fails to generalize to the target domain due to domain shift. \textit{Bottom:} UniDA aligns shared classes across domains while detecting unknown target samples, assigning them to a rejection zone.}

    \label{fig:unida}
\end{figure}

UDA aims to transfer knowledge from a labeled source domain, denoted as $\mathcal{D}^s = \{(x_i^s, y_i^s)\}_{i=1}^{n_s}$, to an unlabeled target domain, $\mathcal{D}^t = \{(x_i^t)\}_{i=1}^{n_t}$. The source domain $\mathcal{D}^s$ follows a joint distribution $\mathcal{P}^s(x^s, y^s)$, while the target domain $\mathcal{D}^t$ follows a different joint distribution $\mathcal{P}^t(x^t, y^t)$, such that $\mathcal{P}^s \neq \mathcal{P}^t$. In UniDA, the domain shift is still considered (i.e., $\mathcal{P}^s \neq \mathcal{P}^t$), but the source and target label sets are not guaranteed to be equal (i.e., $\mathcal{Y}^s \neq \mathcal{Y}^t$). More precisely, one can describe three subsets:

\begin{itemize}  
    \item The \textbf{common label set} (\(\mathcal{Y}^C = \mathcal{Y}^s \cap \mathcal{Y}^t\)), which is the set of shared source and target labels. The target samples of this set should be properly aligned with their corresponding source samples. 
    \item The \textbf{private source label set} (\(\overline{\mathcal{Y}^s} = \mathcal{Y}^s \backslash \mathcal{Y}^t\)), which is the set of labels present in the source domain but not in the target domain. No target samples should be aligned with such source samples so as to avoid mislabeling.   
    \item The \textbf{private target label set} (\(\overline{\mathcal{Y}^t} = \mathcal{Y}^t \backslash \mathcal{Y}^s\)), which is the set of labels present in the target domain but not in the source domain. They form the unknown target samples that must be detected and categorized as unknown.
\end{itemize}  

Given these subsets, it is clear that Universal Domain Adaptation involves two key tasks: i) aligning common samples and ii) detecting unknown target samples. The UniDA problem is illustrated in Fig. \ref{fig:unida}.

\begin{figure}[t]
    \centering
    \includegraphics[width=\linewidth]{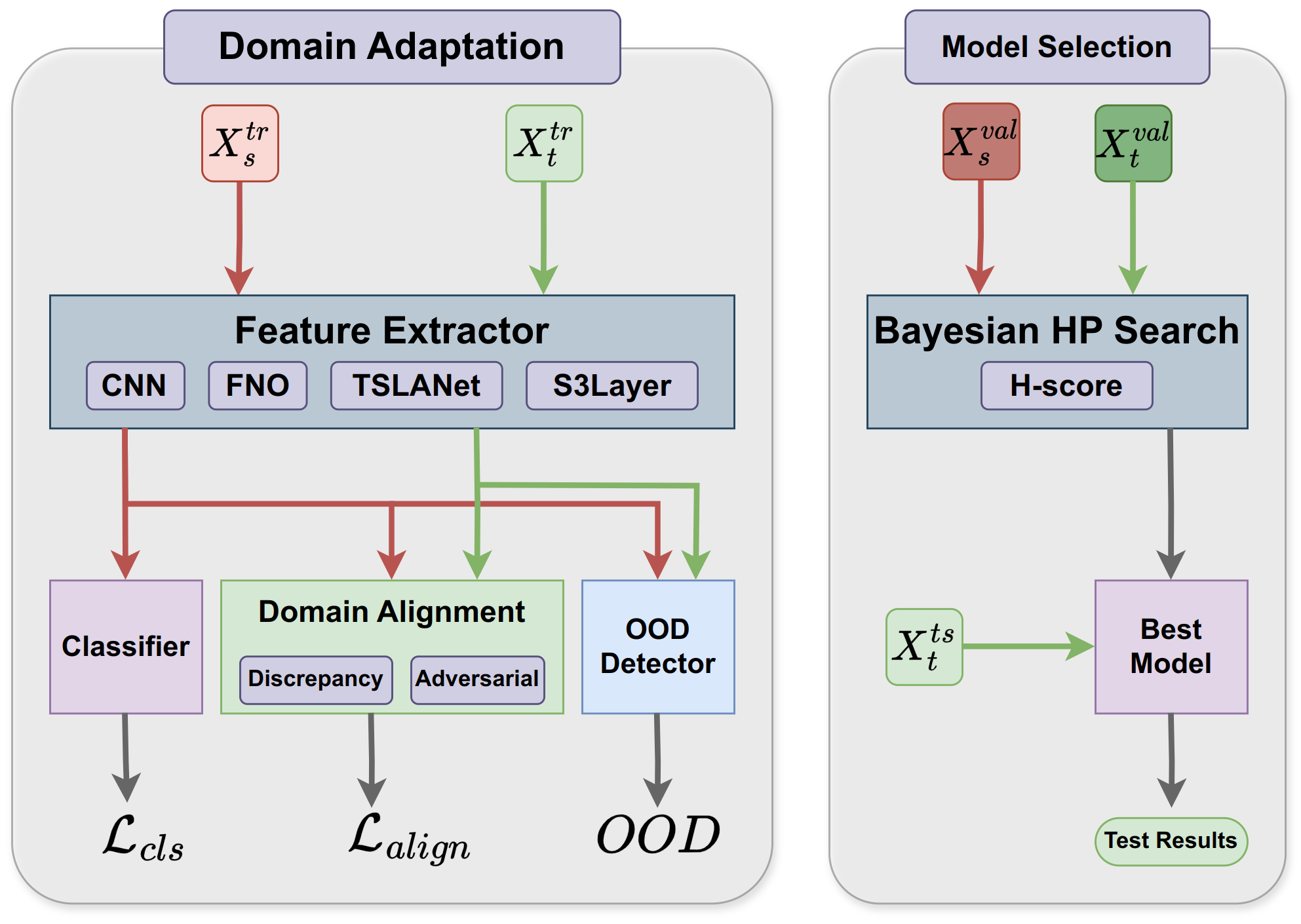}
    \caption{\textbf{Overview of the benchmark:} \textit{Domain Adaptation:} UniDA models extract features using various architectures, learn the task with a classification loss ($\mathcal{L}_{cls}$), align domains via discrepancy/adversarial methods, and detect unknown samples with an OOD detector. \textit{Model Selection:} Hyperparameters are optimized using Bayesian search and H-score to select the best model for final evaluation.}
    \label{fig:flowchart}
\end{figure}

Solving the UDA problem using deep learning involves a feature extractor $g$ and a classifier $f$, trained with two losses: (i) a classification loss, typically cross-entropy, over the labeled source domain, $\mathcal{L}_{\text{ce}}$, and (ii) an alignment loss, $\mathcal{L}_{\text{align}}$, ensuring feature consistency across domains. The latter can be adversarial or discrepancy-based (e.g., MMD, Wasserstein). In UniDA, an additional unknown target sample detector is required, often based on thresholding entropy or other distribution-based distance metrics. This is partially illustrated in Fig. \ref{fig:flowchart}.

\section{Benchmark} \label{III}

An overview of the benchmark is proposed in Fig. \ref{fig:flowchart}. This section describes the studied backbones, models and datasets over the UniDA framework.

\subsection{Model Selection Protocol}

\begin{algorithm}
\caption{Model Selection Procedure}
\label{alg:model_selection}
\begin{algorithmic}[1]
    \Require Number of runs $N_r$, number of validation scenarios $N_{val}$, number of test scenarios $N_{eval}$
    \Require Dataset $\mathbf{D} = \{ \mathcal{D}^1, \mathcal{D}^2, \cdots \mathcal{D}^d\}$
    
    \State $\mathcal{S}_{val}, \mathcal{S}_{eval}  \gets \text{select}(\mathbf{D}, N_{val}), \text{select}(\mathbf{D}, N_{eval})$
    
    \State Initialize $\mathbf{scores} \gets [\,]$ \hfill \textit{\#Vector to store H-scores}
    
    \For{$n \gets 1$ to $N_r$}
        \For{each pair $\{\mathcal{D}^s, \mathcal{D}^t\} \in \mathcal{S}_{val}$}
            \State Split $\mathcal{D}^t$ into $\{\mathcal{D}^t_{tr}, \mathcal{D}^t_{ts}\}$
            \State Train model using $(\mathcal{D}^s, \mathcal{D}^t_{tr})$
            \State $h \gets \text{compute\_H\_score}(\mathcal{D}^t_{ts})$
            \State $\text{append}(\mathbf{scores}, h)$
        \EndFor
    \EndFor
    
    \State \Return $\arg\max\limits_{i}~ \mathbf{scores}_i$ \hfill \textit{\#Select best models}
\end{algorithmic}
\end{algorithm}

UDA datasets generally contain multiple domains, allowing models to be tested across various adaptation scenarios. Formally, a dataset is defined as $\mathbf{D} = \{ \mathcal{D}^1, \mathcal{D}^2, \dots, \mathcal{D}^d \}$, where $d$ represents the number of domains. Each domain contains both data and labels ($\{x_i, y_i\}_{i=0}^n$), as it can be used as either a source or a target domain.  
By sampling domain pairs, we construct UniDA scenarios $S^{s,t} = \{\mathcal{D}^s, \mathcal{D}^t\}$, where $\mathcal{D}^s$ serves as the source domain and $\mathcal{D}^t$ as the target domain. The set of all possible scenarios, $\mathcal{S}$, is then used to form two subsets: $\mathcal{S}_{val}$ for hyperparameter selection and $\mathcal{S}_{eval}$ for final evaluation.

On a given scenario, UniDA models are typically evaluated using the H-score, defined as $(2A_C A_U) / (A_C + A_U)$, where $A_C$ and $A_U$ denote accuracy on common and unknown classes of the target domain \cite{CMU}. While alternative UDA metrics exist that do not require target labels, they fail to assess a model’s performance when new classes appear. We will rely on the mean H-score on the set of validation scenarios for hyperparameter selection.

In the case of the target domain, the target dataset will be split into $\mathcal{D}^t_{tr}$ to train the model and $\mathcal{D}^t_{ts}$ to compute the H-score after training. Remember that the model is not allowed to use the target's labels during training. However, in order to measure the H-score over the target domains, labels are provided for this part of the dataset.

Optimal hyperparameters are determined via hyperparameter search using Bayesian optimization over $N_r$ training runs, selecting the configuration that maximizes the average H-score across $\mathcal{S}_{val}$. The best model is subsequently evaluated on $\mathcal{S}_{eval}$. The protocol is described in Algorithm \ref{alg:model_selection}. This approach facilitates a systematic comparison of the UniDA models across the TS-oriented architectures in this study.

\subsection{Backbones for Time Series}

In this work, we aim at providing a benchmark to determine if TS-oriented architectures present better performance for UniDA tasks. We selected the classical {1D-CNN} that has already demonstrated interesting results on UDA \cite{ADATIME}. In addition, we selected three recent state-of-the-art TS backbones: \textbf{FNO}, \textbf{TSLANet} and \textbf{S3}.

\textbf{FNO} is inspired by the Fourier Neural Operator (FNO) \cite{fno_li2020fourier}. This layer follows a three-step process: (i) applying the Fast Fourier Transform (FFT) to project the input into the frequency domain, (ii) performing convolution operations on the transformed frequency features, and (iii) applying the inverse FFT (iFFT) to return to the original space. A refinement for UDA was introduced in \cite{RAINCOAT}, where a cosine smoothing function is applied to mitigate misalignment caused by noisy frequency components. Instead of directly applying an iFFT, the method extracts the polar coordinates of the frequency coefficients. Typically, a CNN is employed to extract features in the original space, which are then concatenated with frequency-domain features, thereby capturing both time and frequency domain representations. Throughout this paper, we refer to this architecture as FNO.

\textbf{S3} is a Segment, Shuffle, and Stitch (S3) layer \cite{s3_layer} designed to learn an optimal temporal arrangement of TS data. It follows a three-step process: (i) \emph{Segment}: the TS is partitioned into $n$ segments. (ii) \emph{Shuffle}: a priority vector $p \in \mathbb{R}^n$ is learned, where the highest value $p_i$ indicates greater importance for segment $i$. The segments are then reordered accordingly. (iii) \emph{Stitch}: the reordered segments are combined, and the original input is added back to the sequence to retain essential temporal information. By dynamically reordering TS segments, the S3 layer captures temporal dependencies beyond their original chronological order, improving representation learning for TS tasks.

\textbf{TSLANet} \cite{tslanet} relies on two key components: an Adaptive Spectral Block (ASB) and an Interactive Convolutional Block (ICB). Similar to FNO, the ASB operates in the frequency domain using the FFT. However, it utilizes learnable filters along with a dynamic threshold to attenuate high-frequency components before reconstructing the time-domain signal using iFFT. The ICB, on the other hand, employs 1D convolutions to process the extracted features. The output of the ICB is given by:
$$\mathbf{O}_{ICB} = \operatorname{Conv3}(A_1 + A_2)$$
such that:
$$
A_1 = \phi\left(\operatorname{Conv1}(S')\right) \odot \operatorname{Conv2}(S'),
$$
$$
A_2 = \phi\left(\operatorname{Conv2}(S')\right) \odot \operatorname{Conv1}(S').
$$
Here, \(\phi\) represents the GELU activation function and $S'$ the output of an ASB block. The final TSLANet layer $l$ can be expressed as
$$l = \operatorname{ICB}\bigl(\operatorname{LayerNorm}(\operatorname{ASB}(S_{PE}))\bigr).$$
Here, \(S_{PE}\) denotes the augmented patch formed by combining the original patched TS with a positional embedding.

\subsection{Models}
 
Most existing UniDA models have been designed primarily for computer vision tasks. However, by simply replacing the feature extractor, these methods can be applied to TS data as well. To investigate the impact of TS-specific architectures in UniDA, we benchmark five UniDA methods:

\begin{itemize}
    \item \textbf{UAN} \cite{UDA} employs an adversarial domain discriminator to provide source-target alignment by encouraging domain-invariant representations. Unknown sample detection is performed by thresholding the entropy of a non-adversarial domain classifier.
    \item \textbf{OVANet} \cite{OVANET} introduces a one-versus-all (OVA) classification strategy, where multiple binary classifiers are trained, each responsible for distinguishing one known class from the rest. Unknown samples are rejected when classified negatively by all binary classifiers. Source-target alignment is encouraged via entropy minimization.
    \item \textbf{DANCE} \cite{DANCE} adopts a clustering-based alignment approach. It forms target clusters based on similarity and aligns them with source clusters. Unknown sample detection is performed using thresholding over an entropy-based margin to identify unknown samples.
    \item \textbf{PPOT} \cite{PPOT} leverages an OT-based distance metric to align source class prototypes with target samples. Unknown sample detection is performed using a confidence-based threshold over the classifier’s outputs.
    \item \textbf{UniOT} \cite{UniOT} also utilizes OT for alignment but differs in its detection mechanism. It employs the OT transport matrix to identify unknown samples via a fixed threshold. Like PPOT, alignment is achieved through OT-based optimization.
    \item \textbf{UniJDOT} \cite{UniJDOT} uses OT for alignment but relies on an adaptive thresholding approach on the classifier's outputs, regularized by a distance-based metric over the feature space.
\end{itemize}

\subsection{Datasets}

To create UniDA scenarios, the common practice is to remove the labels of one or more classes from both the source and target domains. We study the TS datasets benchmarked in \cite{ADATIME} for UDA: HAR, HHAR, SleepEDF (EDF), WISDM, and Boiler for the UniDA task. The WISDM dataset contains too few samples, while the Boiler dataset contains few classes, making both datasets unsuitable for the UniDA task. The remaining datasets are HAR, HHAR, and EDF. They are all closed-set with 5 to 6 classes, whereas UniDA requires at least one private class per domain. This is solved by removing exactly one class in each domain for these three datasets.

HAR and HHAR are Human Activity Recognition datasets with six activity classes, such as walking, sitting, and biking, collected via multiple sensors worn by human subjects. Both use non-overlapping 128-time-step windows. HAR contains data from 9 channels and HHAR from 3, making them multivariate. Meanwhile, EDF is a sleep stage classification dataset with five sleep stages, using electroencephalography (EEG) recordings from 20 patients.

\section{Experiments} \label{IV}

\begin{table}[t]
\centering
\caption{H-scores (\%) for HAR, HHAR, and EDF}
\resizebox{\linewidth}{!}{
\begin{tabular}{l l c c c c}
\toprule
\multirow{3}{*}{\textbf{Datasets}} & \multirow{3}{*}{\textbf{Methods}} & \multicolumn{4}{c}{\textbf{Backbones}} \\
\cmidrule(lr){3-6}
& & \textbf{CNN} & \textbf{FNO} & \textbf{TSLANet} & \textbf{S3}\\
\midrule 
\multirow{6}{*}{\textbf{HAR}} & \textbf{UAN} & \underline{60.1} & \textbf{62.3} & 32.3 & 52.2 \\
& \textbf{OVANet} & 24.9 & \underline{31.2} & \textbf{32.0} & 27.3 \\
& \textbf{PPOT} & \underline{37.6} & \textbf{62.1} & 6.9 & 35.9 \\
& \textbf{DANCE} & \underline{53.0} & \textbf{54.1} & 2.0 & 46.7 \\
& \textbf{UniOT} & 47.7 & \underline{49.1} & \textbf{53.5} & 37.6 \\
& \textbf{UniJDOT} & \underline{61.0} & \textbf{64.6} & 59.7 & 54.1 \\
\midrule
\multirow{6}{*}{\textbf{HHAR}} & \textbf{UAN} & \textbf{47.8} & 41.1 & 34.0 & \underline{41.9} \\
& \textbf{OVANet} & 27.0 & \textbf{43.7} & \underline{28.3} & 23.3 \\
& \textbf{PPOT} & \underline{40.6} & \textbf{48.6} & 2.1 & 39.4 \\
& \textbf{DANCE} & 40.6 & \underline{41.8} & 0.0 & \textbf{42.4} \\
& \textbf{UniOT} & \underline{51.0} & \textbf{54.3} & 39.9 & 44.6 \\
& \textbf{UniJDOT} & 56.6 & \textbf{61.2} & 55.7 & \underline{57.8} \\
\midrule
\multirow{6}{*}{\textbf{EDF}} & \textbf{UAN} & \underline{54.2} & \textbf{57.6} & 30.2 & 52.1 \\
& \textbf{OVANet} & \textbf{55.4} & 42.2 & 23.5 & \underline{52.7} \\
& \textbf{PPOT} & \textbf{45.0} & \underline{36.9} & 3.8 & 36.8 \\
& \textbf{DANCE} & \textbf{51.9} & 41.4 & 19.2 & \underline{50.6} \\
& \textbf{UniOT} & \textbf{41.4} & \underline{41.2} & 32.3 & 37.7 \\
& \textbf{UniJDOT} & 44.3 & \textbf{55.6} & \underline{55.0} & 50.4 \\
\bottomrule
\end{tabular}
}
\label{tab:hscore_merged}
\end{table}

\setlength{\tabcolsep}{3pt} 
\begin{table}[t]
\centering
\caption{  H-scores (\%) for HAR}
\resizebox{\linewidth}{!}{
\begin{tabular}{ l c c c c c c}
\toprule
\textbf{Scenario} & \textbf{UAN}$^\star$ & \textbf{OVANet}$^\dagger$ & \textbf{PPOT}$^\star$ & \textbf{DANCE}$^\star$ & \textbf{UniOT}$^\dagger$ & \textbf{UniJDOT}$^\star$\\
\midrule 
$12 \rightarrow 16$ & \underline{56.8} & 41.8 & \textbf{61.0} & 49.9 & 48.9 & 50.6 \\
$13 \rightarrow 3$ & 67.3 & 16.4 & \textbf{78.0} & 71.4 & 64.4 & \underline{76.3} \\
$15 \rightarrow 21$ & \underline{78.4} & 53.7 & 36.8 & 66.6 & 53.2 & \textbf{81.5} \\
$17 \rightarrow 29$ & 63.5 & 32.0 & \underline{72.8} & 65.3 & 52.9 & \textbf{78.3} \\
$1 \rightarrow 14$ & \underline{63.8} & 33.5 & 62.8 & \textbf{64.8} & 58.2 & 39.2 \\
$22 \rightarrow 4$ & 63.0 & 36.7 & \underline{68.4} & 41.2 & 54.0 & \textbf{72.4} \\
$24 \rightarrow 8$ & \underline{50.7} & 40.8 & \textbf{58.2} & 38.6 & \underline{50.7} & 47.0 \\
$30 \rightarrow 20$ & 47.2 & 9.2 & \textbf{54.0} & 49.4 & 51.9 & \underline{53.4} \\
$6 \rightarrow 23$ & 66.6 & 12.4 & \underline{75.8} & 68.5 & 52.8 & \textbf{78.7} \\
$9 \rightarrow 18$ & \underline{65.4} & 43.8 & 53.2 & 25.4 & 48.5 & \textbf{68.6} \\
\hdashline
Mean & \underline{62.3} & 32.0 & \underline{62.1} & 54.1 & 53.5 & \textbf{64.6} \\

\bottomrule
\multicolumn{7}{l}{$^\star$ Models trained with FNO $^\dagger$ Models trained with TSLANet}
\end{tabular}
}
\label{tab:hscore_HAR2}
\end{table}

\subsection{Experimental Settings}

Each model is trained for 20 epochs, as additional epochs generally degrade performance \cite{ADATIME}. Backbone hyperparameters are fixed. The CNN backbone consists of three convolutional blocks. The S3 backbone includes three S3 layers followed by a CNN backbone, while the FNO backbone integrates one FNO layer concatenated with CNN outputs. For TSLANet, we use the original backbone from \cite{tslanet}.

To validate the hyperparameters, we select five validation scenarios ($N_{val} = 5$) for HAR and three for HHAR and EDF ($N_{val} = 3$), as the latter datasets contain more samples per domain, increasing computation time. The hyperparameter search is performed over $N_r = 100$ runs, thus testing 100 sets of hyperparameters for each model. Final results are computed over 10 test scenarios ($N_{ts} = 10$). Each model is trained with 10 different seeds per scenario to enhance robustness. 

\subsection{Results}

Table \ref{tab:hscore_merged} presents the average H-score over 10 test scenarios for each dataset, evaluating multiple state-of-the-art UniDA methods combined with various TS-oriented backbones. Table \ref{tab:hscore_HAR2} shows more detailed results for the HAR dataset. 
Among the backbones, a simple CNN consistently delivers strong performance across most cases, closely followed by FNO.
Since FNO concatenates its Fourier-based layer outputs with those of a CNN backbone, its performance is expected to be comparable or superior to CNN. In contrast, TSLANet performs poorly, rarely achieving the highest H-score and ranking second only three times. Similarly, S3 fails to outperform CNN or FNO. Since S3 layers are followed—not concatenated—by a CNN backbone, its poorer performance suggests information loss before reaching the CNN. A similar issue applies to TSLANet, where the convolutive ICB block follows the spectral ASB block. These results indicate that, despite their state-of-the-art performance in classification, forecasting, and anomaly detection, TSLANet and S3 lack robustness. They fail to provide the domain generalization required for UniDA, making them more sensitive to domain shifts compared to CNN-based layers. Regarding the tested methods, PPOT and DANCE are highly dependent on the backbone, consistently achieving low H-scores across all datasets. In contrast, UniJDOT achieves the best performance on HAR and HHAR and ranks second on EDF. It also demonstrates high robustness, delivering competitive results across architectures and datasets. Most methods frequently register H-scores below 50\%, highlighting their instability, whereas UniJDOT remains above this threshold in all but one case—scoring 44.3\%.
These results underscore the critical role of backbone selection in UniDA performance, with FNO and CNN emerging as the most effective choices, while S3 and TSLANet lack robustness. UniJDOT stands out as the most robust method across datasets and architectures. Its stability is attributed to distance-based regularization and adaptive thresholding for unknown sample detection.

\section{Conclusion} \label{V}

This paper introduced a standardized protocol for evaluating UniDA models on TS and presented the first benchmark focusing on TS-specific feature extractors. Our results highlight the critical role of architecture selection, with CNN and FNO outperforming newer models like TSLANet and S3. UniJDOT proved to be highly robust, consistently achieving strong results across all backbones. Moreover, the failure of recent architectures to consistently improve performance suggests that state-of-the-art representation learning, despite excelling in classification and anomaly detection, may be ineffective for UniDA. These findings emphasize the need for more accurate, tailored TS backbone architectures in UniDA.

\section*{Acknowledgment}

The authors acknowledge the support of the French Agence Nationale de la Recherche (ANR), under grant ANR-23-CE23-0004 (project ODD) and of the STIC AmSud project DD-AnDet under grant N°51743NC.

\bibliographystyle{ieeetr}
\bibliography{main}
\vspace{12pt}

\end{document}